\documentclass[11pt,a4paper]{article}
\usepackage[hyperref]{conll-2019}
\usepackage{times}
\usepackage{latexsym}

\usepackage{url}
\usepackage{graphicx}
\usepackage{amsmath}
\usepackage{booktabs}
\usepackage{algorithm}
\usepackage{algorithmic}
\usepackage{multirow}
\usepackage{caption}
\usepackage{subcaption}
\usepackage{tabu}
\usepackage{makecell}
\usepackage{footnote}
\usepackage{footmisc}

\aclfinalcopy 

\title{A Dual-Attention Hierarchical Recurrent Neural Network\\ for Dialogue Act Classification}

\author{Ruizhe Li\textsuperscript{$\spadesuit$}, Chenghua Lin\textsuperscript{$\heartsuit$}, Matthew Collinson\textsuperscript{$\spadesuit$}, Xiao Li\textsuperscript{$\spadesuit$} and Guanyi Chen\textsuperscript{$\clubsuit$}\\
\textsuperscript{$\spadesuit$}Department of Computing Science, University of Aberdeen, UK\\
\texttt{\{r02rl17, matthew.collinson, x.li\}@abdn.ac.uk}\\
\textsuperscript{$\heartsuit$}Department of Computer Science, University of Sheffield, UK\\
\texttt{c.lin@sheffield.ac.uk}\\
\textsuperscript{$\clubsuit$}Department of Information and Computing Sciences, Utrecht University, The Netherlands \\ 
\texttt{g.chen@uu.nl}}

\date{}

\begin{document}
\maketitle
\begin{abstract}
Recognising dialogue acts (DA) is important for many natural language processing tasks such as dialogue generation and intention recognition. In this paper, we propose a dual-attention hierarchical recurrent neural network for DA classification. Our model is partially inspired by the observation that conversational utterances are normally associated with both a DA and a topic, where the former captures the social act and the latter describes the subject matter. However, such a dependency between DAs and topics has not been utilised by most existing systems for DA classification. With a novel dual task-specific attention mechanism, our model is able, for utterances, to capture information about both DAs and topics, as well as information about the interactions between them. Experimental results show that by modelling topic as an auxiliary task, our model can significantly improve DA classification, yielding better or comparable performance to the state-of-the-art method on three public datasets.

\end{abstract}

\section{Introduction}
Dialogue Acts (DA) are semantic labels of utterances, which are
crucial to understanding communication: much of a speaker's intent is expressed, explicitly or implicitly, via social actions (e.g., questions or requests) associated with utterances~\cite{searle1969speech}.
Recognising DA labels is important for many natural language processing tasks. For instance, in dialogue systems, knowing the DA label of an utterance supports its interpretation as well as the generation of an appropriate response~\cite{searle1969speech,chen2018dialogue}. 
In the security domain, being able to detect intention in conversational texts can effectively support the recognition of sensitive information exchanged in emails or other communication channels, which is critical to timely security intervention~\cite{verma2012detecting}.

A wide range of techniques have been investigated for DA classification. Early works on DA classification are mostly based on general machine learning techniques, framing the problem either as multi-class classification (e.g., using  SVMs~\cite{liu2006using} and dynamic Bayesian networks~\cite{dielmann2008recognition}) or a structured prediction task (e.g., using Conditional Random Fields~\cite[CRF]{kim2010classifying,chen2018dialogue,raheja2019dialogue}).
Recent studies to the problem of DA classification have seen an increasing uptake of deep learning techniques, where promising results have been obtained. 
Deep learning approaches typically model the dependency between adjacent utterances~\cite{ji2016latent,lee2016sequential}. Some researchers further account for dependencies among both consecutive utterances and consecutive DAs, i.e., both are considered factors that influence natural dialogue~\cite{kumar2017dialogue,chen2018dialogue}. 
There is also work exploring different deep learning architectures (e.g., hierarchical CNN or RNN/LSTM) for incorporating context information for DA classification~\cite{liu2017using}. 

It has been observed that conversational utterances are normally associated with both a DA and a topic, where the former captures the social act (e.g., promising) and the latter describes the subject matter~\cite{wallace2013generative}. 
It is also recognised that the types of DA associated with a conversation are likely to be influenced by the topic of the conversation~\cite{searle1969speech,wallace2013generative}. For instance, conversations relating to topics about \textit{customer service} might be more frequently associated with DAs of type Wh-question (e.g., \textit{Why my mobile is not working?}) and a complaining statement~\cite{bhuiyan2018don}; whereas meetings covering administrative topics about resource allocation are likely to exhibit significantly more defending statements and floor grabbers (e.g., \textit{Well I mean -  is the handheld really any better?})~\cite{wrede2003relationship}. 
However, such a reasonable source of information, surprisingly, has not been explored in the deep learning literature for DA classification. 
We assume that modelling the topics of utterances as additional contextual information
may effectively support DA classification. 

In this paper, we propose a dual-attention hierarchical recurrent neural network with a CRF (DAH-CRF) for DA classification. 
Our model is able to account for rich context information with the developed dual-attention mechanism, which, 
in addition to accounting for the dependencies between utterances, can further capture, for utterances, information about both topics and DAs. Topic is a useful source of context information which has not previously been explored in existing deep learning models for DA classification. 
Second, compared to the flat structure employed by existing models~\cite{khanpour2016dialogue,ji2016latent}, our hierarchical recurrent neural network can represent the input at the character, word, utterance, and conversation levels, preserving the natural hierarchical structure of a conversation. To capture the topic information of conversations, we propose a simple automatic utterance-level topic labelling mechanism based on LDA~\cite{blei2003latent}, which avoids expensive human annotation and improves the generalisability of our model. 

We evaluate our model against several strong baselines~\cite{wallace2013generative,ji2016latent,kumar2017dialogue,chen2018dialogue,raheja2019dialogue} on the task of DA classification. 
Extensive experiments conducted on three public datasets (i.e., Switchboard Dialog Act Corpus (SWDA), DailyDialog (DyDA), and the Meeting Recorder Dialogue Act corpus (MRDA)) show that by modelling the topic information of utterances as an auxiliary task, our model can significantly improve DA classification for all datasets compared to a base model without modelling topic information. Our model also 
yields better or comparable performance to state-of-the-art deep learning method~\cite{raheja2019dialogue} in classification accuracy. 

To summarise, the contributions of our paper are three-fold:  
(1) we propose to leverage topic information of utterances, a useful source of contextual information  which has not previously been explored in existing deep learning models for DA classification; 
(2) we propose a dual-attention hierarchical recurrent neural network with a CRF which respects the natural hierarchical structure of a conversation, and is able to incorporate rich context information for DA classification, achieving better or comparable performance to the state-of-the-art; 
(3) we develop a simple topic labelling mechanism, showing that using the automatically acquired topic information for utterances can effectively improve DA classification.  

\section{Related Work} \label{sec:relate}

Broadly speaking, methods for DA classification can be divided into two  categories: multi-class classification (e.g., 
SVMs~\cite{liu2006using} and dynamic Bayesian networks~\cite{dielmann2008recognition}) and structured prediction tasks including HMM~\cite{stolcke2000dialogue} and CRF~\cite{kim2010classifying}. 
Recently, deep learning has been widely applied in many NLP tasks, including DA classification. \citet{kalchbrenner2013recurrent}  proposed to model a DA sequence with a 
RNN where sentence representations were constructed by means of a convolutional neural network (CNN). \citet{lee2016sequential}  tackled DA classification with a model built upon RNNs and CNNs. Specifically, their model can leverage the information of preceding texts, which can effectively help improve the DA classification accuracy. 
A latent variable recurrent neural network was developed for jointly modelling sequences of words and discourse relations between adjacent sentences~\cite{ji2016latent}. In their work, the shallow discourse structure is represented as a latent variable and the contextual information from preceding utterances are modelled with a RNN. 

\citet{kumar2017dialogue}  proposed a hierarchical
Bi-LSTM model with a CRF for DA classification, where the inter-utterance and intra-utterance information are encoded by a hierarchical Bi-LSTM and the dependency between DA labels is captured by a CRF. \citet{chen2018dialogue} developed a CRF-Attentive Structured Network (CRF-ASN) for DA classification. They applied structured attention network to the CRF layer in order to model contextual utterances and corresponding DAs together. \citet{raheja2019dialogue} achieved the state-of-the-art performance on the SWDA dataset by employing a self-attention mechanism, a CRF layer and character-level embeddings.

In addition to modelling dependency between utterances, various contexts have also been explored for improving DA classification or joint modelling DA under multi-task learning. 
For instance, \citet{wallace2013generative}  proposed a generative joint sequential model to classify both DA and topics of patient-doctor conversations. Their model is similar to the factorial LDA model~\cite{paul2012factorial}, which generalises LDA to assign each token a $K$-dimensional vector of latent variables. We would like to emphasise that the model of~\citet{wallace2013generative}, 
only assumed that each utterance is generated conditioned on the previous and current topic/DA pairs. In contrast, our model is able to model the dependencies of all preceding  utterances of a conversation, and hence can better capture the effect between DAs and topics.

\section{Methodology}


\begin{figure*}[t]
  \centering
  \includegraphics[scale=0.4]{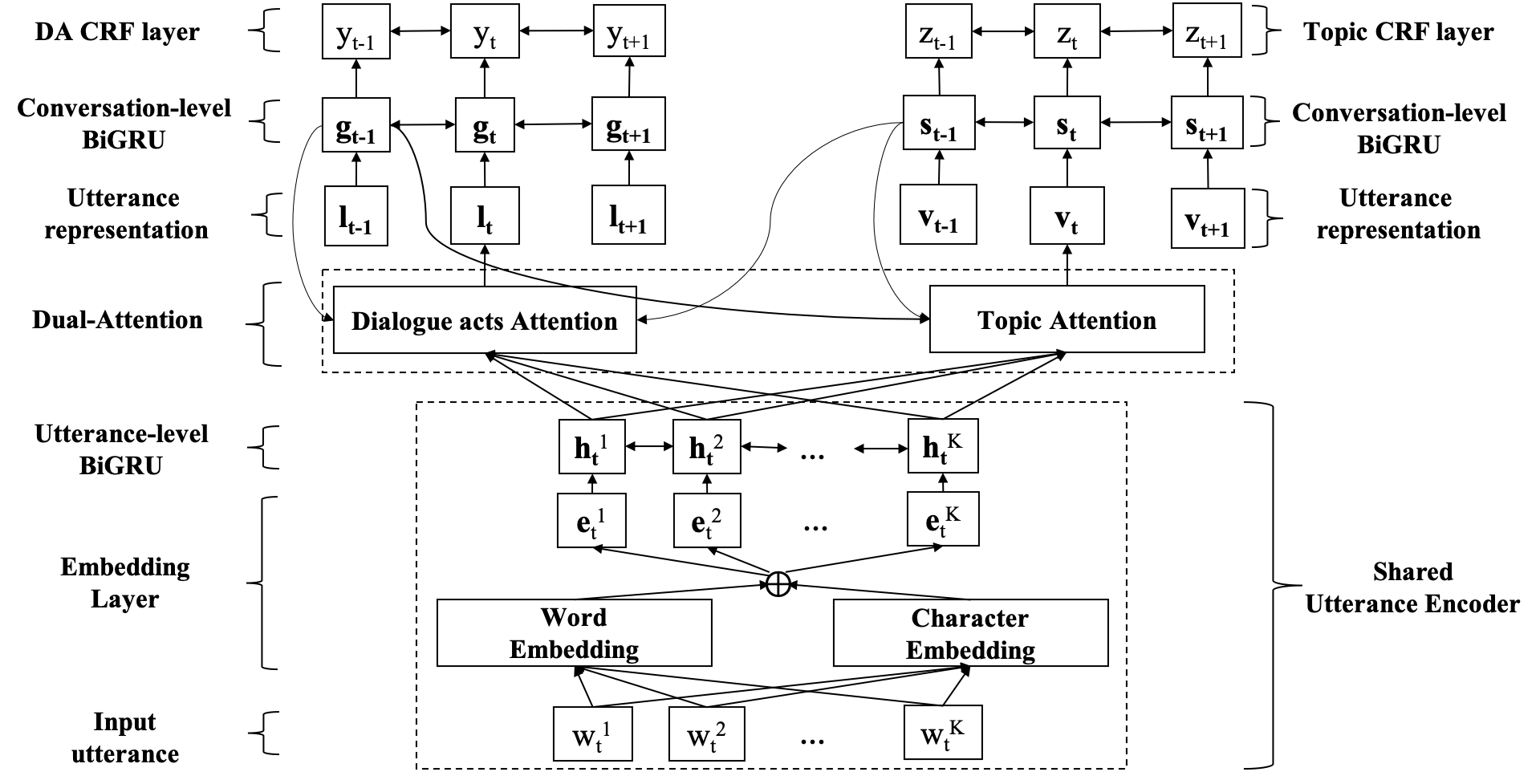}
  \caption{Overview of the dual-attention hierarchical recurrent neural network with a CRF.}
  \label{fig:model}
\end{figure*}

Given a training corpus $\mathcal{D}=\langle(C_n, Y_n, Z_n)\rangle_{n=1}^N$, where $C_n = \langle u_{t}^n \rangle_{t=1}^T$ is a conversation containing a sequence of $T$ utterances, $Y_n = \langle y_{t}^n \rangle_{t=1}^T$ and $Z_n = \langle z_{t}^n \rangle_{t=1}^T$ are the corresponding labels of DA and topics for $C_n$, respectively. Each utterance $u_{t} = \langle w_t^i \rangle_{i=1}^K$ of $C_n$ is a sequence of $K$ words. Our goal is to learn a model from $\mathcal{D}$, such that, given an unseen conversation $C_u$, the model can predict the DA labels 
of the utterances of $C_u$. 

Figure~\ref{fig:model} gives an overview of the proposed Dual-Attention Hierarchical recurrent neural network with a CRF (DAH-CRF). 
A shared utterance encoder encodes each word $w_t^i$ of an utterance $u_t$ into a vector $\mathbf{h}_t^i$.
The DA attention and topic attention mechanisms capture DA and topic information as well as the interactions between them. 
The outputs of the dual-attention are then encoded in the conversation-level sequence taggers (i.e., $\mathbf{g} _t$ and $\mathbf{s}_t$), based on the corresponding utterance representations (i.e., $\mathbf{l} _t$ and $\mathbf{v}_t$). Finally, the target labels (i.e., $y _{t}$ and $z_{t}$) are predicted in the CRF layer. 

\subsection{Shared Utterance Encoder}
In our model, we adopt a shared utterance encoder to encode the input utterances.  Such a design is based on the rationale that the shared encoder can transfer parameters between two tasks and reduce the risk of overfitting~\cite{ruder2017overview}. 
Specifically, the shared utterance encoder is implemented using the bidirectional gated recurrent unit~\cite[BiGRU]{cho2014properties}, which encodes each utterance $u_{t} = \langle w_t^i \rangle_{i=1}^K $ of a conversation $C_{n}$ as a series of hidden states $\langle \mathbf{h}_t^i\rangle_{i=1}^K$. Here, $i$ indicates the timestamp of a sequence, and  we define $\mathbf{h}_t^i$ as follows 
\begin{equation} \label{eq:1}
\mathbf { h }_t^i = {  \overrightarrow { \mathbf { h } } _ t^i \oplus \overleftarrow  { \mathbf { h } } _ t^i }
\end{equation}
where $\oplus$ is an operation for concatenating two vectors, and $\overrightarrow { \mathbf { h } } _ t^i$ and $\overleftarrow  { \mathbf { h } } _ t^i$ are the $i$-th hidden state of the forward gated recurrent unit~\cite[GRU]{cho2014properties} and backward GRU for $w_t^i$, respectively. Formally, the forward GRU $\overrightarrow { \mathbf { h } } _ t^i$ is computed as follows 
\def\GRU{\mathop{\rm GRU}}
\begin{equation} \label{eq:2}
    \overrightarrow { \mathbf { h } } _ t^i = \GRU (\overrightarrow { \mathbf { h } } _ t^{i-1},\mathbf{e} _ t^i)
\end{equation}
where $\mathbf{e} _ t^i$ is the concatenation of the word embedding and the character embedding of word $w_t^i$. Finally, the backward GRU encodes \textbf{$u_t$} from the reverse direction (i.e. $w_t ^K  \rightarrow  w_t ^1$) and generates $\langle \overleftarrow{\mathbf{h}_t^i}\rangle_{i=1}^K$ following the same formulation as the forward GRU.

\subsection{Task-specific Attention} 

Recall that one of the key challenges of our model is to capture for each utterance, information about both DAs and topics, as well as information about the interactions between them. We address this challenge by incorporating into our model a novel task-specific dual-attention mechanism, which accounts for both DA and topic information extracted from utterances.   
In addition, DAs and topics are semantically relevant to  different words in an utterance. With the proposed attention mechanism, our model can also assign different weights to the words of an utterance by learning the degree of importance of the words to the DA or topic labelling task, i.e., promoting the words which are important to the task and reducing the noise introduced by less important words.

For each utterance $u_t$, the DA attention calculates a weight vector $\langle \alpha_t^i\rangle_{i=1}^K$ for $\langle\mathbf{h}_t^i\rangle_{i=1}^K$, the hidden states of $u_t$. $u_t$ can then be represented as an attention vector $\mathbf{l}_t$ computed as follows 
\begin{equation} \label{eq:3}
\mathbf{l} _ { t } = \sum _ { i=1 }^{K} \alpha _ { t } ^ { i } \mathbf{h} _ { t } ^ { i }
\end{equation}

In contrast to the traditional attention mechanism~\cite{bahdanau2014neural}, which only depends on one set of hidden vectors from the Seq2Seq decoder, the DA attention of our model relies on two sets of hidden vectors, i.e., $\mathbf{g}_{t-1}$ of the conversation-level DA tagger and $\mathbf{s}_{t-1}$ of the conversation-level topic tagger, where dual attention mechanism can capture, for utterances, information about both DAs and topics as well as the interaction between them. Specifically, the weights $\langle \alpha_t^i\rangle_{i=1}^K$ for the DA attention are calculated as follows: 
\def\softmax{\mathop{\rm softmax}}
\begin{gather} \label{eq:4}
\alpha_t^i = \softmax (o_t^i)\\
    \resizebox{0.95\hsize}{!}{%
$o _ { t } ^ { i } =\mathbf{w}^{\top}_a\tanh \left( \mathbf{W} ^ { ( \mathrm { act } ) } (\mathbf{s} _ { t-1 }  \oplus \mathbf { g } _ { t - 1 }  \oplus \mathbf{h}_{t}^{i}) + \mathbf { b } ^ { ( \mathrm { act } ) } \right)$%
}
\end{gather} 

The topic attention layer has a similar architecture to the DA attention layer, which takes as input both $\mathbf{s}_{t-1}$ and $\mathbf{g}_{t-1}$. 
The weight vector $\langle \beta_t ^i \rangle_{i=1} ^K$ for the  topic attention output $\mathbf{v}_t$ can be calculated similar to Eq. \ref{eq:3} and Eq. \ref{eq:4}.
Note that $\mathbf{w}_a$, 
$\mathbf{W} ^ { ( \mathrm{ act } )}$, 
and $\mathbf { b } ^ { ( \mathrm { act } ) }$
are vectors of parameters that  need to be learned during training. 

\subsection{Conversational Sequence Tagger}

\noindent\textbf{CRF sequence tagger for DA.}~~~The conversational CRF sequence tagger for DA predicts the next DA $y_t$ conditioned on the conversational hidden state $\mathbf{g}_t$ and adjacent DAs (c.f.~Figure~\ref{fig:model}). Formally, this conditional probability of the whole conversation can be formulated as

\begin{equation}
    p\left( y _ { 1 : T} | C ; \theta \right) =
\frac { \prod _ { t = 1 } ^ { T } \Psi \left( y _ { t - 1 } , y _ { t } , \mathbf{g} _ { t } ; \theta \right) } { \sum _ { Y } \prod _ { t = 1 } ^ { T } \Psi \left( y _ { t - 1 } , y _ { t } , \mathbf{g} _ { t } ; \theta \right) }
\end{equation}
\begin{equation}
    \begin{split}
        \Psi \left( y _ { t - 1 } , y _ { t } , \mathbf{g} _ { t } ; \theta \right) &= \Psi _ {emi} \left( y _ { t },\mathbf{g} _ { t } \right) \Psi_{tran} \left( y _ { t - 1 },y _ { t } \right)\\
&= \mathbf{g} _ { t } \left[ y _ { t } \right] \mathbf {P} _ { y _ { t } , y _ { t - 1 } }
    \end{split}
\end{equation}
Here the feature function $\Psi(\cdot)$ includes two score potentials: emission and transition. The emission potential $\Psi_ {emi}$ regards utterance representation $\mathbf{g}_t$ as the unary feature. The transition potential $\Psi_{tran}$ is a pairwise feature constructed from a $T \times T$ state transition matrix $\mathbf {P}$, where $T$ is the number of DA classes, and  $\mathbf {P}_{y_{t}, y_{t-1}}$ is the probability of transiting from state $y_{t-1}$ to $y_{t}$.  $C = \langle u_{t} \rangle_{t=1}^T$ is the sequence of all utterances seen so far, $\theta$ is the parameters of the CRF layer. $\mathbf{g}_t$ is calculated in a BiGRU similar to Eq.~\ref{eq:1} and Eq.~\ref{eq:2}:
\begin{gather} \label{eq:8}
\mathbf { g }_t = {  \overrightarrow { \mathbf { g } } _ t \oplus \overleftarrow  { \mathbf { g } } _ t }\\
    \overrightarrow { \mathbf { g } } _ t = \GRU (\overrightarrow { \mathbf { g } } _ {t-1},\mathbf{l} _ t)%
\end{gather}

\noindent\textbf{CRF sequence tagger for topic.}~~~The conversational CRF sequence tagger for topic is designed to predict topic $z_t$ conditioned on $\mathbf{v}_t$ and adjacent topics, which can be calculated similar to the formulation of the CRF tagger for DA.

\noindent\textbf{Training the model.}~~~Let $\Theta$ be all the model parameters that need to be estimated for DAH-CRF.  $\Theta$ then is estimated based on $\mathcal{D}=\langle(C_n, Y_n, Z_n)\rangle_{n=1}^N$ (i.e., a corpus with $N$ conversations) by maximising the following objective function
\begin{align}
    \mathcal{L} = \sum_{n=1} ^ { N }
&\left[\log \left( p \left( y _{1:T}^n | C_n;\Theta \right) \right) \right.
    \nonumber \\ 
 & 
    \left.
 		\qquad +  \alpha \log\left( p \left( z _{1:T}^n  | C_n;\Theta \right)\right) \right]
\end{align}
The hyper-parameter $\alpha $ controls the contribution of the conversational topic tagger towards the objective function. In our experiments, $\alpha = 0.5$ is determined using the validation datasets. During the test, the optimal DA or topic sequence is calculated using the Viterbi algorithm~\cite{viterbi1967error}.
\begin{equation}
    Y' = \arg \max _ { y_{1:T} \in Y } p ( y_{1:T} | C , \Theta )
\end{equation}

\subsection{Automatically Acquiring Topic Labels}

To avoid expensive human annotation and to improve the generalisability of our model, we propose to label the topic of each utterance of the datasets using LDA~\cite{blei2003latent}. 
While perplexity has been widely used for model selection for LDA~\cite{chenghua2011probabilistic,he2012online}, we employ a topic coherence measure proposed by~\cite{roder2015exploring} to determine the optimal topic number for each dataset, which combines the indirect cosine measure with the normalised pointwise mutual information~\cite[NPMI]{bouma2009normalized} and the Boolean sliding window. Empirically, we found the latter yields much better topic clusters than perplexity for supporting DA classification. 

\begin{figure}[tb]
  \centering
  \includegraphics[scale=0.5]{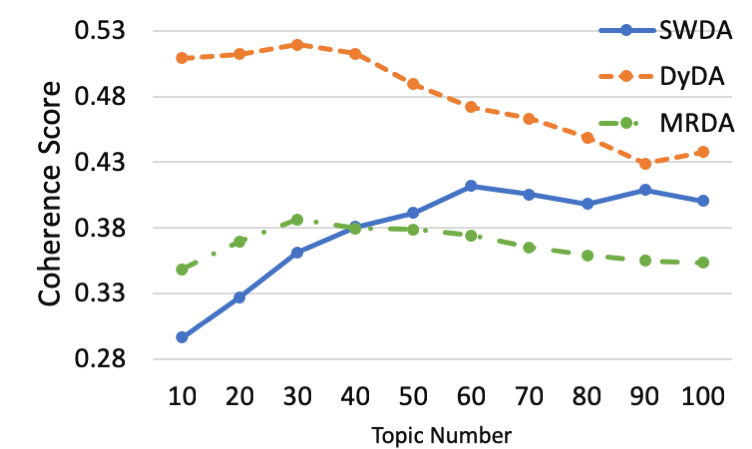}
  \caption{Coherence score of LDA on three datasets.}
  \label{fig:co_per_lda}
\end{figure}
 

We treat each conversation as a document and train topic models using Gensim with topic number settings ranging from 10 to 100 (using an increment step of 10).  Gibbs sampling is used to estimate the model posterior and for each model we run 1,000 iterations. For each trained model, we calculate the averaged coherence score of the extracted topics using Gensim\footnote{\url{https://radimrehurek.com/gensim/models/coherencemodel.html}}, an implementation following~\cite{roder2015exploring}. Figure~\ref{fig:co_per_lda} shows the topic coherence score for each topic number setting for all datasets, from which we determine that the optimal topic number setting for SWDA, DyDA, and MRDA are 60, 30, and 30, respectively. 

Based on the optimal models (i.e., a trained LDA model using the optimal topic number setting), we assign topic labels to the datasets with two different strategies, i.e., conversation-level labelling ($conv$) and utterance-level labelling ($utt$). For conversation-level labelling, we assign the topic label with the highest marginal probability to the conversation based on the corresponding per-document topic proportion estimated by LDA. Every utterance of the conversation then shares the same topic label of the conversation. For utterance-level labelling, there is an additional step to perform inference on every utterance based on corresponding optimal model (e.g., for every utterance of SWDA, we do inference using the LDA trained on SWDA with 60 topics), and assign the topic label with the highest marginal probability to the utterance. Therefore, the topic labels of the utterances of the same conversation could be different for utterance-level labelling.

\begin{table}[tb]
\resizebox{\columnwidth}{!}{%
  \centering
  \begin{tabular}{lcccccc}
    \Xhline{2.5\arrayrulewidth}
    Dataset & $|C|$ & $|T|$& $|V|$ & Training & Validation & Testing\\
    \hline
    SWDA   & 42 & 66 & 20K & 1003/193K & 112/23K & 19/5K\\
    DyDA   & 4  & 10 & 22K & 11K/92.7K   & 1K/8.5K   & 1K/8.2K\\
    MRDA   & 5  & - & 15K & 51/77.9K   &  11/15.8K  &  11/15.5K\\
    \Xhline{2.5\arrayrulewidth}
  \end{tabular}
  }
  \caption{$|C|$ is the number of DA classes, $|T|$ is the number of 
  manually labelled conversation-level topic classes, $|V|$ is the vocabulary size. Training, Validation and Testing indicate the number of conversations/utterances in the respective splits.}
  \label{T:datasets}
\end{table}

\section{Experimental Settings}

\subsection{Datasets}
We evaluate the performance of our model on three public DA datasets with different characteristics, namely, Switchboard Dialog Act Corpus~\cite[SWDA]{jurafsky1997switchboard}, Dailydialog~\cite[DyDA]{li2017dailydialog}, and the Meeting Recorder Dialogue Act corpus~\cite[MRDA]{janin2003icsi}. 

\noindent \textbf{SWDA}\footnote{\url{https://web.stanford.edu/~jurafsky/ws97/manual.august1.html}} consists of 1,155 two-sided telephone conversations manually labelled with 66 conversation-level topics (e.g., \textit{taxes}, \textit{music}, etc.) and 42 utterance-level DAs (e.g., \textit{statement-opinion}, \textit{statement-non-opinion}, \textit{wh-question}).

\noindent \textbf{DyDA}\footnote{\url{http://yanran.li/dailydialog}} contains 13,118 human-written daily conversations, manually labelled with 10 conversation-level topics (e.g., \textit{tourism}, \textit{politics}, \textit{finance}) as well as four utterance-level DA classes, i.e., \textit{inform}, \textit{question}, \textit{directive} and \textit{commissive}. The former two classes are information transfer acts,  while the latter two are action discussion acts.

\noindent \textbf{MRDA}\footnote{\url{http://www1.icsi.berkeley.edu/~ees/dadb/}}  contains 75 meeting conversations annotated with 5 DAs, i.e., Statement (S), Question (Q), Floorgrabber (F), Backchannel (B), and Disruption (D). The average number of utterances per conversation is 1,496. There are no manually annotated topic labels available for this dataset. 

\subsection{Implementation Details}
For all experimental datasets, the top 85\% highest frequency words were indexed. For SWDA and MRDA, we split training/validation/testing datasets following~\cite{stolcke2000dialogue,lee2016sequential}.
For DyDA, we  used the standard split from the original dataset~\cite{li2017dailydialog}.
The statistics of the experimental datasets are summarised in Table~\ref{T:datasets}. 
We represented input data with 300-dimensional Glove word embeddings~\cite{pennington2014glove} and 50-dimensional character embeddings~\cite{ma2016end}. 
We set the dimension of the hidden layers (i.e., $\mathbf { h }_t^i$, $\mathbf{g}_t$ and $\mathbf{s}_t$) to 256 and applied a dropout layer to both the shared encoder and the sequence tagger at a rate of 0.2. The Adam optimiser~\cite{kingma2014adam} was used for training with an initial learning rate of 0.001 and a weight decay of 0.0001. 
Each utterance in a mini-batch was padded to the maximum length for that batch, and the maximum batch-size allowed was 50.

\subsection{Baselines}
We compare the proposed DAH-CRF model incorporating utterance-level topic labels extracted by LDA (denoted as DAH-CRF+LDA$_{utt}$) against five strong baselines and two variants of our own models:\\
\textbf{JAS}\footnote{\url{https://github.com/bwallace/JAS}}: A generative joint, additive, sequential model of topics and speech acts in patient-doctor communication~\cite{wallace2013generative};\\
\textbf{DRLM-Cond}\footnote{\url{https://github.com/jiyfeng/drlm}}: A latent variable recurrent neural network for DA classification~\cite{ji2016latent};\\
\textbf{Bi-LSTM-CRF}\footnote{\url{https://github.com/YanWenqiang/HBLSTM-CRF}}: A hierarchical Bi-LSTM with a CRF 
to classify DAs~\cite{kumar2017dialogue};\\
\textbf{CRF-ASN}: An attentive structured network with a CRF for DA classification~\cite{chen2018dialogue};\\
\textbf{SelfAtt-CRF}: A hierarchical Bi-GRU with self-attention and CRF~\cite{raheja2019dialogue};\\
\textbf{DAH-CRF+MANUAL$_{conv}$}: Use the manually annotated conversation-level topic labels (i.e., each utterance of the conversation shares the same topic) for DAH-CRF model training rather than the topic labels automatically acquired from LDA;\\
\textbf{DAH-CRF+LDA$_{conv}$}: Use conversation-level topic labels automatically acquired from LDA for DAH-CRF model training.

Note that only JAS (a non-deep-learning model) has attempted to model both DAs and topics, whereas all the deep learning baselines do not model topic information as a source of context for DA classification. All the baselines mentioned above use the same test dataset as our models for all experimental datasets.

\section{Experimental Results}

\begin{table}[tb] 
\resizebox{\columnwidth}{!}{%
  \centering 
  \begin{tabular}{c|lccc} 
    \Xhline{2.5\arrayrulewidth}
    &\textbf{Model} & \textbf{SWDA} & \textbf{MRDA} & \textbf{DyDA}\\
     \hline
    \multirow{4}{*}{\rotatebox[origin=c]{90}{Baselines}} & JAS  & 71.2 & 81.3 & 75.9 \\
    & DRLM-Cond & {~~77.0}$^\dag$ & 88.4 & 81.1 \\
    & Bi-LSTM-CRF & {~~79.2}$^\dag$ & {~~90.9}$^\dag$ & 83.6 \\
    & CRF-ASN & {~~80.8}$^\dag$ & {~~91.4}$^\dag$ & - \\
    & SelfAtt-CRF &{~~\textbf{82.9}}$^\dag$ &{~~91.1}$^\dag$ & - \\\hline
    \multirow{4}{*}{\rotatebox[origin=c]{90}{Ours}} & DAH-CRF + MANUAL$_{conv}$  & 80.9 & - & 86.5  \\
    & DAH-CRF + LDA$_{conv}$ & 80.7 & 91.2 & 86.4 \\ 
    & DAH-CRF + LDA$_{utt}$ & 82.3 & \textbf{92.2} & \textbf{88.1} \\ \hline
    & Human Agreement & 84.0 & - & - \\
    \Xhline{2.5\arrayrulewidth}
  \end{tabular}
  }
  \caption{DA classification accuracy.  
  $^\dag$ indicates the results which are reported from the prior publications.}
  \label{T:swda}
\end{table}

\subsection{Dialogue Acts Classification}
Table~\ref{T:swda} shows the DA classification accuracy of our models and the baselines on three experimental datasets. We fine-tuned the model parameters for JAS, DRLM-Cond and Bi-LSTM-CRF in order to make the comparison as fair as possible. The implementation of CRF-ASN and SelfAtt-CRF are not available so we can only report their results for SWDA and MRDA based on the original papers~\cite{chen2018dialogue,raheja2019dialogue}.

It can be observed that by jointly modelling DA and topics,
DAH-CRF+LDA$_{utt}$ outperforms the two best baseline models  SelfAtt-CRF and CRF-ASN  around 1\% on the MRDA dataset.
Our model also gives similar performance to SelfAtt-CRF, the baseline which achieved the state-of-the-art performance on the SWDA dataset (i.e., 82.3\% vs. 82.9\%). 
While both manually annotated and automatically acquired topic labels are effective, we see that DAH-CRF+LDA$_{utt}$  outperforms both  DAH-CRF+MANUAL$_{conv}$ and DAH-CRF+LDA$_{conv}$, i.e., with over 1.6\% gain on  DyDA and over 1.4\% on SWDA  (significant; paired t-test $p<.01$). It is also observed that DAH-CRF+MANUAL$_{conv}$ and DAH-CRF+LDA$_{conv}$ perform very similar to each other.

\begin{table}[tb] 
\resizebox{\columnwidth}{!}{%
  \centering 
  \begin{tabular}{>{\raggedright}p{4cm}ccc} 
    \Xhline{2.5\arrayrulewidth}
    \textbf{Model} & \textbf{SWDA} & \textbf{MRDA} & \textbf{DyDA}\\
     \hline
    SAH & 76.2 & 88.5 & 82.5\\
     SAH-CRF & 78.4 & 89.6 & 84.1 \\
    DAH + LDA$_{utt}$ & 79.5 & 91.1 & 86.0 \\
    DAH-CRF + LDA$_{utt}$ (without Dual-Att)& 81.0 & 91.3 & 86.3 \\
    DAH-CRF + LDA$_{utt}$ & 82.3 & 92.2 & 88.1 \\
    
    \Xhline{2.5\arrayrulewidth}
  \end{tabular}
  }
  \caption{Ablation studies of DA classification.}
  \label{T:ablation}
\end{table}

\subsection{Ablation Study Results}
We conducted ablation studies (see Table~\ref{T:ablation}) in order to evaluate the contribution of the components of our DAH-CRF+LDA$_{utt}$ model, and more importantly, the effectiveness of leveraging topic information for supporting DA classification.

DAH-CRF+LDA$_{utt}$ (without Dual-Att) removes the dual-attention component from DAH-CRF+LDA$_{utt}$, and   DAH+LDA$_{utt}$ removes the CRF from DAH-CRF+LDA$_{utt}$ but retaining the dual-attention component. SAH is a Single-Attention Hierarchical RNN model without a CRF, i.e., a simplified version of DAH+LDA$_{utt}$ that only models DAs with topical information omitted. As can be seen in Table~\ref{T:ablation},  DAH+LDA$_{utt}$ achieves over 3\% averaged gain on all datasets when compared to SAH, which clearly shows that leveraging topic information can effectively support DA classification. It is also observed that both the dual-attention mechanism and the CRF component are beneficial, but are more effective on the SWDA and DyDA datasets than MRDA.

In summary, while all the analysed model components are beneficial, the biggest   gain is obtained by jointly modelling DAs and topics.

\begin{figure}[tb]
  \centering
  \includegraphics[scale=0.45]{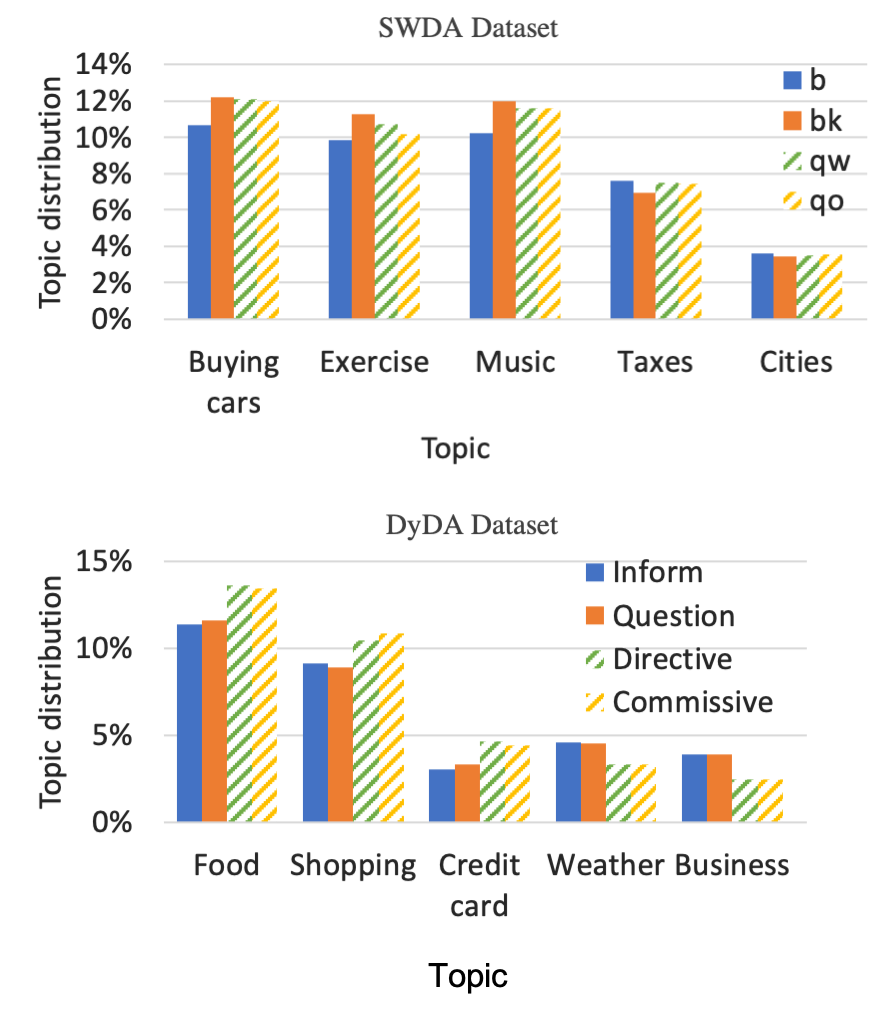}
  \caption{We highlight the prominent topics for some example DAs. The topic distribution of a topic $k$ under a DA label $d$ is calculated by averaging the marginal probability of topic $k$ for all utterances with the DA label $d$.}
  \label{fig:swda_dyda_topic_dist}
\end{figure}

\subsection{Analysing the Effectiveness of Joint Modelling Dialogue Act and Topic}

\begin{figure*}[tb]
    \centering
        \includegraphics[scale=0.49]{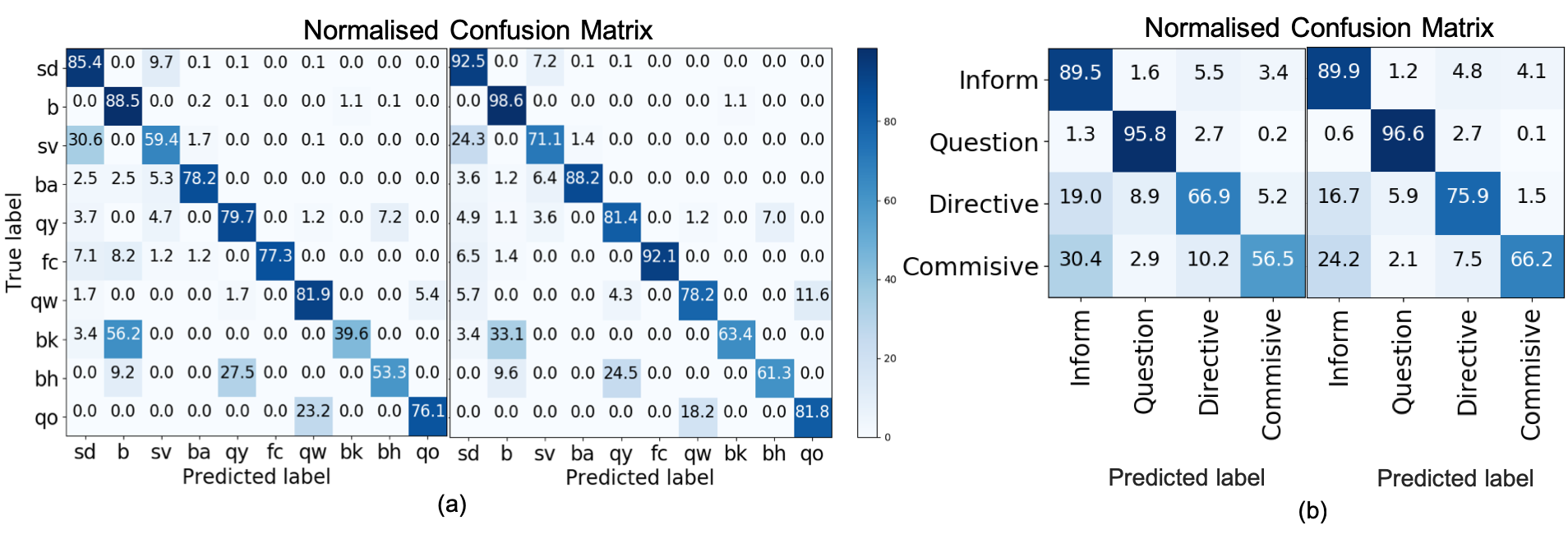}
    \caption{The normalized confusion matrix of DAs using SAH-CRF (left) and DAH-CRF+LDA$_{utt}$ (right) on SWDA (a) and DyDA (b).}
    \label{fig:swda_cm}
\end{figure*}

\begin{figure*}[tb]
  \centering
  \includegraphics[scale=0.35]{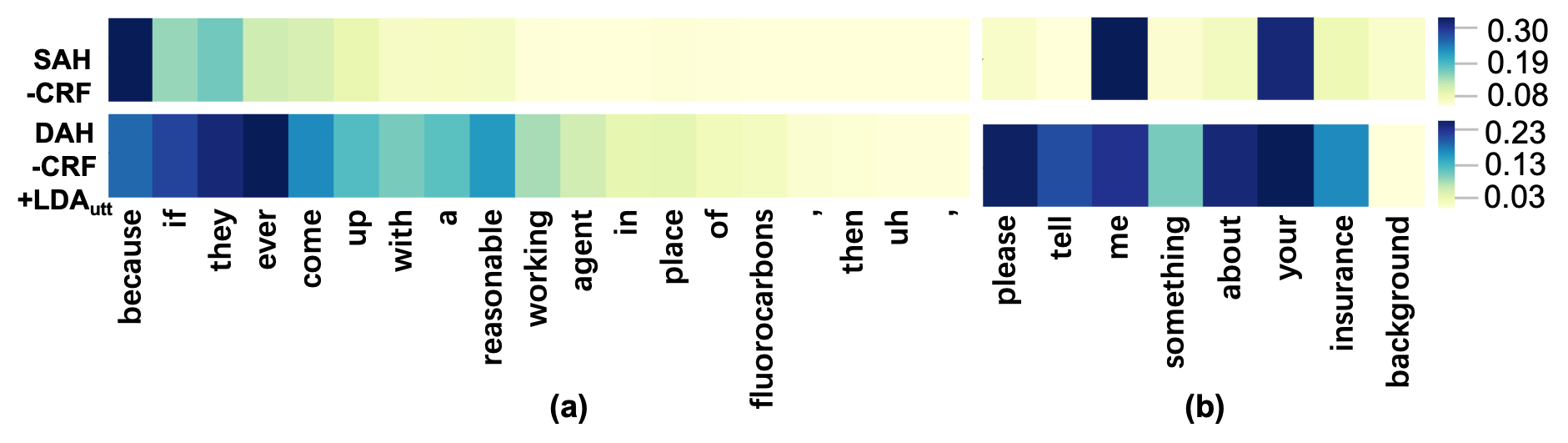}
  \caption{DA Attention visualisation using SAH-CRF and DAH-CRF+LDA$_{utt}$ on (a) SWDA and (b) DyDA datasets. The true labels of the utterances above are \textit{sd} (\textit{statement-non-opinion}) and \textit{Directive}, respectively. SAH-CRF misclassified the  DA as \textit{sv} (\textit{statement-opinion}) and \textit{Inform} whereas  DAH-CRF+LDA$_{utt}$ gives correct prediction for both cases.}
  \label{fig:swda_dyda_attention}
\end{figure*}

In this section, we provide detailed analysis on why DAH-CRF+LDA$_{utt}$ can yield better performance than  SAH-CRF by jointly modelling DAs and topics. Due to the page limit, our discussion focuses on SWDA and DyDA datasets.

Figure~\ref{fig:swda_cm} shows the normalized confusion matrix derived from 10 DA classes of SWDA for both SAH-CRF and DAH-CRF+LDA$_{utt}$ models. 
It can be observed that DAH-CRF+LDA$_{utt}$ yields improvement on recall for many DA classes compared to SAH-CRF, e.g., 23.8\% improvement on \textit{bk} and 11.7\% on \textit{sv}. For \textit{bk} (\texttt{Response Acknowledge}) which has the highest improvement level, we see that the improvement largely comes from the reduction of misclassifing \textit{bk} to \textit{b} (\texttt{Acknowledge Backchannel}). The key difference between \textit{bk} and \textit{b} is that an utterance labelled with \textit{bk} has to be produced within a question-answer context, whereas \textit{b} is a ``continuer'' simply representing a response to the speaker~\cite{jurafsky1997switchboard}.
It is not surprising that SAH-CRF makes poor prediction on the utterances of these two DAs: they share many syntactic cues, e.g., indicator words such `okay', `oh', and `uh-huh', which can easily confuse the model. When comparing the topic distribution of the utterances under the \textit{bk} and \textit{b} categories (cf. Figure~\ref{fig:swda_dyda_topic_dist}), we found topics relating to personal leisure (e.g., buying cars, music, and exercise) are much more prominent in \textit{bk} than \textit{b}. By leveraging the topic information, DAH-CRF+LDA$_{utt}$ can better handle the confusion cases and hence improve the prediction for \textit{bk} significantly.

There are also cases where DAH-CRF+LDA$_{utt}$ performs worse than SAH-CRF. Take the DA pair of \textit{qo} (\texttt{Open Question}) and \textit{qw} (\texttt{wh-questions}) as an example. \textit{qo} refers to questions like `\textit{How about you?}' and its variations (e.g., `\textit{What do you think?}'), whereas \textit{qw} represents  wh-questions which are much more specific in general  (e.g. `\textit{What other long range goals do you have?'}). 
SAH-CRF gives quite decent performance in distinguishing \textit{qw} and \textit{qo} classes. This is somewhat reasonable,  as linguistically the utterances of these two classes are quite different, i.e., the \textit{qw} utterance expresses very specific question and is relatively lengthy, whereas \textit{qo} utterances tends to be very brief. We see that DAH-CRF+LDA$_{utt}$  performs worse than SAH-CRF: a greater number of \textit{qw} utterances are misclassified by DAH-CRF+LDA$_{utt}$ as \textit{qo}. This might be attributed to the fact that topic distributions of \textit{qw} and \textit{qo} are similar to each other (see  Figure~\ref{fig:swda_dyda_topic_dist}), i.e., incorporating the topic information into DAH-CRF may cause these two DAs to be less distinguishable for the model.

We also conducted a similar analysis on the DyDA dataset. As can be seen from the confusion matrices shown in Figure~\ref{fig:swda_cm}, DAH-CRF+LDA$_{utt}$ gives improvement over SAH-CRF for all the four DA classes of DyDA. In particular, \texttt{Directives} and \texttt{Commissive} achieve higher improvement margin compared to the other two classes, where the improvement are largely attributed to less number of instances of the \texttt{Directives} and \texttt{Commissive} classes being mis-classified into \texttt{Inform} and \texttt{Questions}. 
Examining the topic distributions in Figure~\ref{fig:swda_dyda_topic_dist}
reveals that  \texttt{Directives} and \texttt{Commissive} classes are more relevant to the topics such as \textit{food}, \textit{shopping}, and \textit{credit card}. In contrast, the topics of \texttt{Inform} and \texttt{Questions} classes are more about \textit{business}, and \textit{weather}.

Finally, Figure~\ref{fig:swda_dyda_attention} shows the DA attention visualisation examples of SAH-CRF and DAH-CRF+LDA$_{utt}$ for an utterance from SWDA and DyDA. For SWDA, it can be seen that SAH-CRF gives very high weight to  the word ``because'' and de-emphasizes other words. However, DAH-CRF+LDA$_{utt}$ can capture more important words (e.g.,  ``if'', ``reasonable'', etc.) and correctly predicts the DA label as \textit{sd}. For DyDA, SAH-CRF only focuses on ``me'' and ``your'', but DAH-CRF+LDA$_{utt}$ captures more words relevant to \texttt{Directive}, such as ``please'', ``tell'', etc. To summarise, DAH-CRF+LDA$_{utt}$ can capture  more significant words related to the corresponding DA, by modelling both DAs and topic information with the dual-attention mechanism.

\section{Conclusion}
In this paper, we developed a dual-attention hierarchical recurrent neural network with a CRF for DA classification. 
With the proposed task-specific dual-attention mechanism, our model is able to capture information about both DAs and topics, as well as information about the interactions between them. 
Moreover, our model is generalised by leveraging an unsupervised model to automatically acquire topic labels.
Experimental results based on three public datasets show that  modelling utterance-level topic information as an auxiliary task can effectively improve DA classification, and that our model is able to achieve better or comparable  performance to the state-of-the-art deep learning methods for DA classification. 

We envisage that our idea of modelling topic information for improving DA classification can be adapted to other DNN models, e.g., to encode topic labels into word embeddings and then concatenate with the utterance-level or conversation-level hidden vectors of our baselines, e.g. SelfAtt-CRF. It will also be interesting to explicitly take into account speaker's role in the future.

\section*{Acknowledgment}
This work is supported by the award made by the UK Engineering and Physical Sciences Research Council (Grant number: EP/P011829/1).

\bibliography{conll-2019}
\bibliographystyle{acl_natbib}

\end{document}